\documentclass[11pt,a4paper]{article}
\usepackage[hyperref]{acl2020}
\usepackage{varwidth}
\usepackage{times}
\usepackage{multicol}
\usepackage{booktabs}
\usepackage{multirow}
\usepackage{latexsym}
\usepackage{todonotes}
\usepackage{amsmath}
\usepackage{cleveref}

\usepackage{graphicx}
\usepackage{capt-of}
\usepackage{tabularx}
\usepackage{color,soul}
\usepackage{pbox}

\usepackage{microtype}

\aclfinalcopy 

\title{Generating Rationales in Visual Question Answering}
\author{Hammad A. Ayyubi\textsuperscript{\scriptsize{*}}, Md. Mehrab Tanjim\textsuperscript{\scriptsize{*}}, Julian J. McAuley, and Garrison W. Cottrell\\
Department of Computer Science\\
UC San Diego\\
{\tt\small \{hayyubi, mtanjim, jmcauley, gary\}@eng.ucsd.edu}
}
\begin{document}

\date{}

\maketitle

\begin{abstract}
Despite recent advances in Visual Question Answering (VQA), it remains a challenge to determine how much success can be attributed to sound reasoning and comprehension ability. 
We seek to investigate this question by proposing a new task of \textit{rationale generation}. Essentially, we task a VQA model with 
generating rationales
for the answers it predicts. We use 
data from the
Visual Commonsense Reasoning (VCR) task, as it contains 
ground-truth
rationales along with visual questions and answers. 
We first
investigate commonsense understanding in one of the leading 
VCR models,
ViLBERT, by generating rationales from 
pretrained 
weights using a state-of-the-art language model, GPT-2. 
Next, we seek to
jointly train ViLBERT with GPT-2 in an end-to-end fashion with the dual task of predicting the answer in VQA and generating rationales. We show that this kind of 
training 
injects commonsense understanding in the VQA model
through quantitative and qualitative evaluation metrics. 
\end{abstract}

\section{Introduction}
\label{intro}

Visual Question Answering (VQA) \cite{Agrawal_2016} 
tasks are an important assessment of joint language-vision understanding. To perform well on VQA, a model must 
understand
the given question and then find a relevant answer from the image. 
A great deal of success 
has
been achieved 
in this task with state-of-the-art models \cite{chen2019uniter} achieving high accuracy on challenging VQA datasets \cite{Goyal_2017, ren2015exploring}.

However, a 
critical question 
is how well 
these models
``understand''
the image, questions, and the answers that they are predicting.
Are they just exploiting biases in the questions \cite{sain2018overcoming,Johnson_2017,cadene2019rubi},
images \cite{Agrawal_2018,Goyal_2017} 
or 
the data \cite{Jabri_2016,manjunatha2018explicit}?
Answering these questions can help shed light on the limitations of existing VQA approaches, and could also lead to more interpretable/explainable VQA systems.

It is a non-trivial task to
evaluate a model's simultaneous understanding of the three components (questions, images, and answers). 
Previous work has analyzed  models' understanding of questions \cite{shah2019cycleconsistency,Agrawal_2016_analyze_VQA}, 
images \cite{Das_2017,goyal2016transparent} 
and answers \cite{Chandrasekaran_2018} individually. 
They have done so by perturbing words (language modality) or investigating heat-maps of images (vision modality). A joint measure of question, image and answer understanding requires an approach which can simultaneously understand and test both linguistic and visual modalities.

To address the 
above
challenges, we propose the 
novel 
task\footnote{Source code will be provided at the time of publication} of generating rationales in VQA as a measure of model's comprehensive understanding.
This task not only requires the model to understand the questions (linguistic modality) and the images (visual modality), but it also requires the model to rationalize the predicted answer in relation to the question and image.

As we need gold standard rationales to compare 
to
the generated rationales, we use the dataset provided by the Visual Commonsense Reasoning (VCR) task \cite{zellers2018recognition}. This dataset contains questions, images, 
multiple-choice
answers (four choices) and four options for rationales, out of which one is correct.
We train 
models for this particular
VQA task in this dataset: choose a correct answer from four options, given a question and image. Then, we 
task the model with
generating rationales for the answers they predict. We compare the generated rationale against the 
ground-truth
rationale from the dataset, as a measure of the model's comprehensive ability.

We employ this approach to investigate one of the leading models on the VCR task---ViLBERT \cite{lu2019vilbert}.
Further, we propose a way
to explicitly inject commonsense understanding into the model by jointly training ViLBERT and the language model, GPT-2 \cite{radford2019language} on the multi-task
objective
of 
predicting the answers and generating rationales. The idea is that by backpropagating the loss of rationale generation into the answer prediction model, sound reasoning can be injected to improve  model's comprehensive understanding.



\section{Approach}
\begin{figure}[t]
\includegraphics[width=\linewidth]{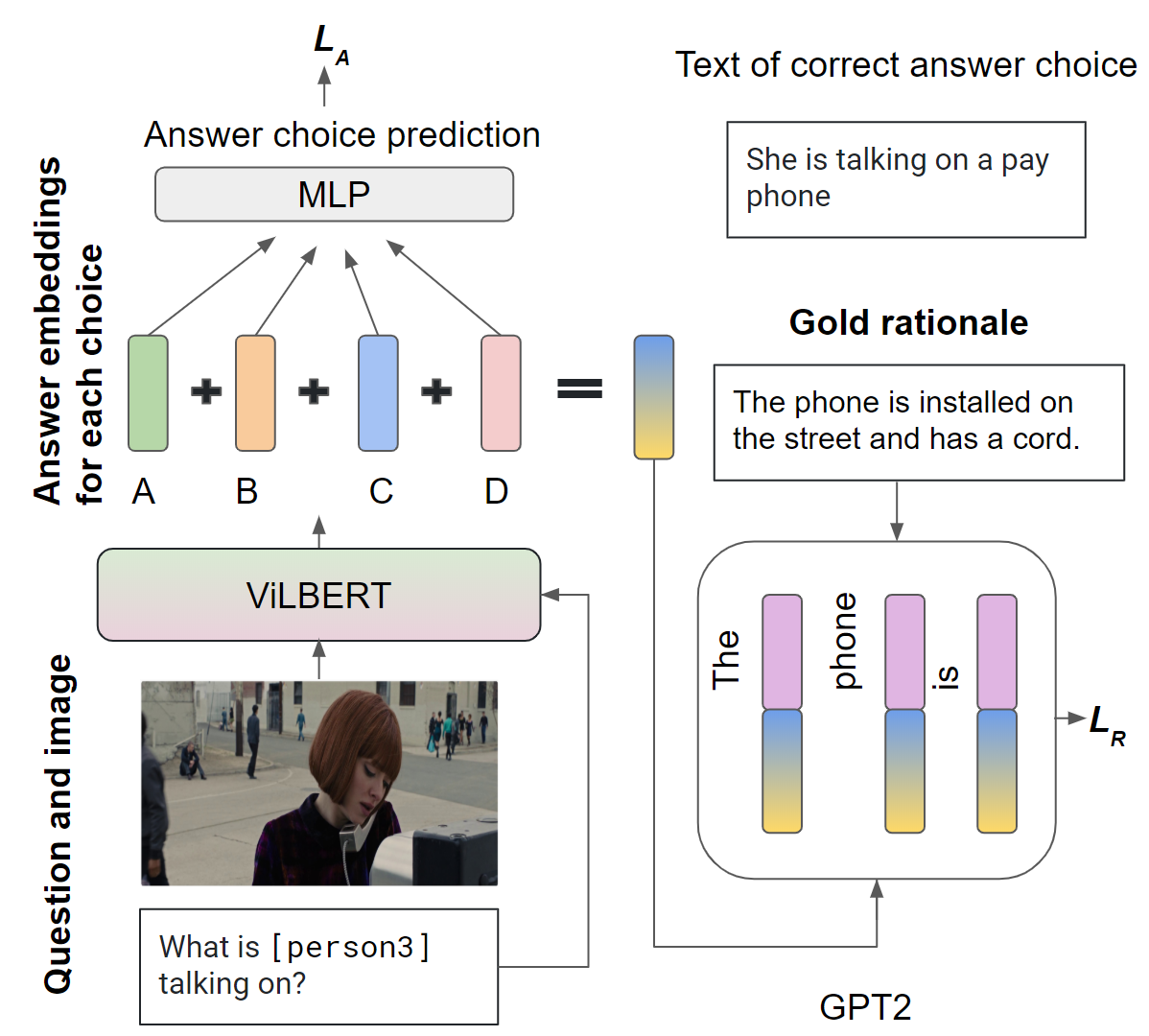}
\caption{High-level overview of the proposed task. We ask the VQA model to generate rationale for the answer it is predicting.}
\label{fig:model}
\end{figure}

The proposed task is illustrated in \Cref{fig:model}.
We approach our task of generating rationale by breaking 
it into two essential components: 
first calculating a predicted answer embedding, $E_{A_p}$, from the VQA model (pretrained ViLBERT)
by
providing it with the given image $I$ and the question $Q$; and second, feeding the predicted answer embedding $E_{A_p}$ to a language model (pretrained GPT2).

\subsection{Predicted Answer Embedding}
Our VQA task
is represented in 
terms of a
question $Q$, an image $I$, and 
four answer choices
$A_1 \ldots A_4$,
among which
the model must choose the correct option.
The models approach this task by outputting an embedding for each answer options, $E_{A_i}$. The four answer embeddings are later passed through a linear-softmax layer to predict answer scores. So, if $f$ is a VQA model (e.g., ViLBERT) parameterized by $\theta$, then the embeddings and softmax scores are calculated as follows:
\begin{gather}
    E_{A_i} = f(Q, I,A_i;\theta) \quad 
    \resizebox{.18\hsize}{!}{
    $ \forall i \in \{1,4\} $
    } \\
    \small{
    s_i = \mathit{Softmax}(\mathit{Linear}(E_{A_i};\theta_l))\quad \resizebox{.18\hsize}{!}{
    $ \forall i \in \{1,4\} $
    } 
    }
    \label{eq:soft_scores}
\end{gather}
The act of choosing the most probable answer embedding out of four options will make the network non-differentiable. To address this issue, we calculate the predicted answer embedding by taking the average of each answer option embedding $E_{A_i}$ weighted by their softmax scores:

\begin{equation}
\label{eq:cal-Ap}
\resizebox{.45 \hsize}{!}{
 $ E_{A_p} = \sum_{i=1}^4 E_{A_i} \times s_i $
 }
\end{equation}
\subsection{Generating Rationales}
We formulate the task of generating rationale as conditional language generation, conditioned on the predicted answer embedding and previously generated rationale tokens. Specifically, 
if $r=r_1,\dots, r_n$ is the rationale, and 
$E_{A_p}$ is the predicted answer embedding,
we maximize the following log likelihood:
\begin{equation}
    \log(P(r)) = \sum_{i=1}^n \log(P(r_i|E_{A_p}, r_1, \ldots, r_{i-1})) \nonumber
\end{equation}
Here, $r_i$ is the $i^{th}$ token of the rationale. The language model is then fine-tuned using the gold standard rationale for 
the corresponding visual question-answer from the VCR dataset.

\section{Experiments}
We use the Visual Commonsense Reasoning (VCR) dataset \cite{zellers2018recognition}
since (in addition to
visual questions and answers) this dataset 
includes rationales. The dataset has 290,000 multiple choice questions derived from 110,000 movie scenes. We report all results on the validation set as the test set labels are not available 
while the VCR challenge is ongoing.

We use 
ViLBERT \cite{lu2019vilbert} as the reference VQA model. For the language model, we use 
the 124 million parameter pretrained GPT-2 (small) \cite{radford2019language}.
We use a batch size of 32, initial learning rate of 2e-5 and train the models for 20 epochs for all our experiments.

\subsection{Evaluating VQA Model Understanding}
\label{eval_vqa}
Since we want to investigate how well the VQA reference model (ViLBERT) already ``understands'' the image, the question and the answer, we freeze the pretrained weights of the model. We extract predicted answer embeddings using
\cref{eq:cal-Ap}, and generate rationales using GPT-2,
conditioned on this answer embedding.
We fine-tune GPT-2 using the ground-truth rationale from the dataset. We call this model ViLBERT-Fr (ViLBERT-Frozen).

\subsection{Injecting Commonsense into VQA}
In this setting, we want to explicitly enforce commonsense understanding in the VQA model \textit{while} predicting the answer. We follow the same procedure as in 
\Cref{eval_vqa}, except 
the weights of ViLBERT are fine-tuned as well. We train ViLBERT with GPT-2 in an end-to-end fashion with the dual loss of answer prediction, $\mathcal{L_A}$ (via a cross-entropy loss) and rationale generation, $\mathcal{L_R}$ (via a causal language modeling loss). The final loss is:
\setlength{\abovedisplayskip}{2pt}
\setlength{\belowdisplayskip}{2pt}
\begin{equation}\label{eq:losses}
    \mathcal{L} = \lambda \mathcal{L_A} + \mathcal{L_R} 
\end{equation}
where $\lambda$ is the weight 
fine-tuned during our experiments.
\begin{table}[t]
\small
\centering
\begin{tabular}{lcc}
\toprule
\textbf{Loss} & \textbf{VQA Accuracy} & \textbf{BLEU-1} \\
\midrule
 $\lambda=1$ & 70.18 & 11.24\\
 $\lambda=3$ & 69.96 & 10.78\\
 $\lambda=10$ & 70.24 & 10.55\\
 $\lambda=1000$ & 69.84 & \textbf{11.27}\\
 var & 70.19 & 11.15\\
 \hline
 kldiv (ViLBERT-Ra) & \textbf{70.45} & \textbf{11.27}\\
\bottomrule
\end{tabular}
\caption{Comparison of Losses. $\lambda$ is from \cref{eq:losses}, var = Homoscedastic Uncertainty loss and kldiv = KL-Divergence regularizer added.}
\label{tab:vqa_results}
\end{table}

\vspace{-0.5em}
\subsection{Evaluation Metric}
We use BLEU \cite{papineni-etal-2002-bleu} and 
ROUGE
\cite{lin-2004-rouge} to compare generated and gold standard rationales. 
In addition to these n-gram metrics, we are also interested in comparing the semantic similarity of rationales. 
We follow \citet{Huang2018ContentBasedIR} 
and calculate sentence embeddings using the InferSent model proposed by \citet{Conneau_2017}, followed by cosine similarity measurement to compare generated rationales with the gold standard.
\begin{table}[t]
\small
\centering
  \begin{tabular}{lcc}
    \toprule
    \multirow{2}{*}[-0.5\dimexpr \aboverulesep + \belowrulesep + \cmidrulewidth]{\textbf{Metrics}}
    & \multicolumn{2}{c}{\textbf{Models}}\\
    \cmidrule(l){2-3}
    & ViLBERT-Fr & \hspace{-2mm}\parbox{21mm}{\centering ViLBERT-Ra \\}\hspace{-2mm}\\
    \midrule
    \textbf{BLEU-1} & 8.92 & \textbf{11.27} \\
    \textbf{BLEU-4} &  0.56  & \textbf{0.68} \\
    \textbf{ROUGE-1} &  13.52 & \textbf{17.08} \\
    \textbf{ROUGE-L} &  11.28 &  \textbf{14.15} \\
    \textbf{Cosine Similarity}\hspace{-6mm} & 0.57 & \textbf{0.60} \\
    \textbf{VQA Accuracy}\hspace{-6mm} & 69.58 & \textbf{70.45} \\
    \bottomrule
  \end{tabular}
  \caption{Comparison of generated rationale vs gold standard rationale on the validation set of VCR dataset.
  }
  \label{tab:results}
\end{table}
\begin{table*}[ht]
  \begin{minipage}[b]{0.4\linewidth}
  \vspace{0pt}
    \centering
    \includegraphics[width=60mm]{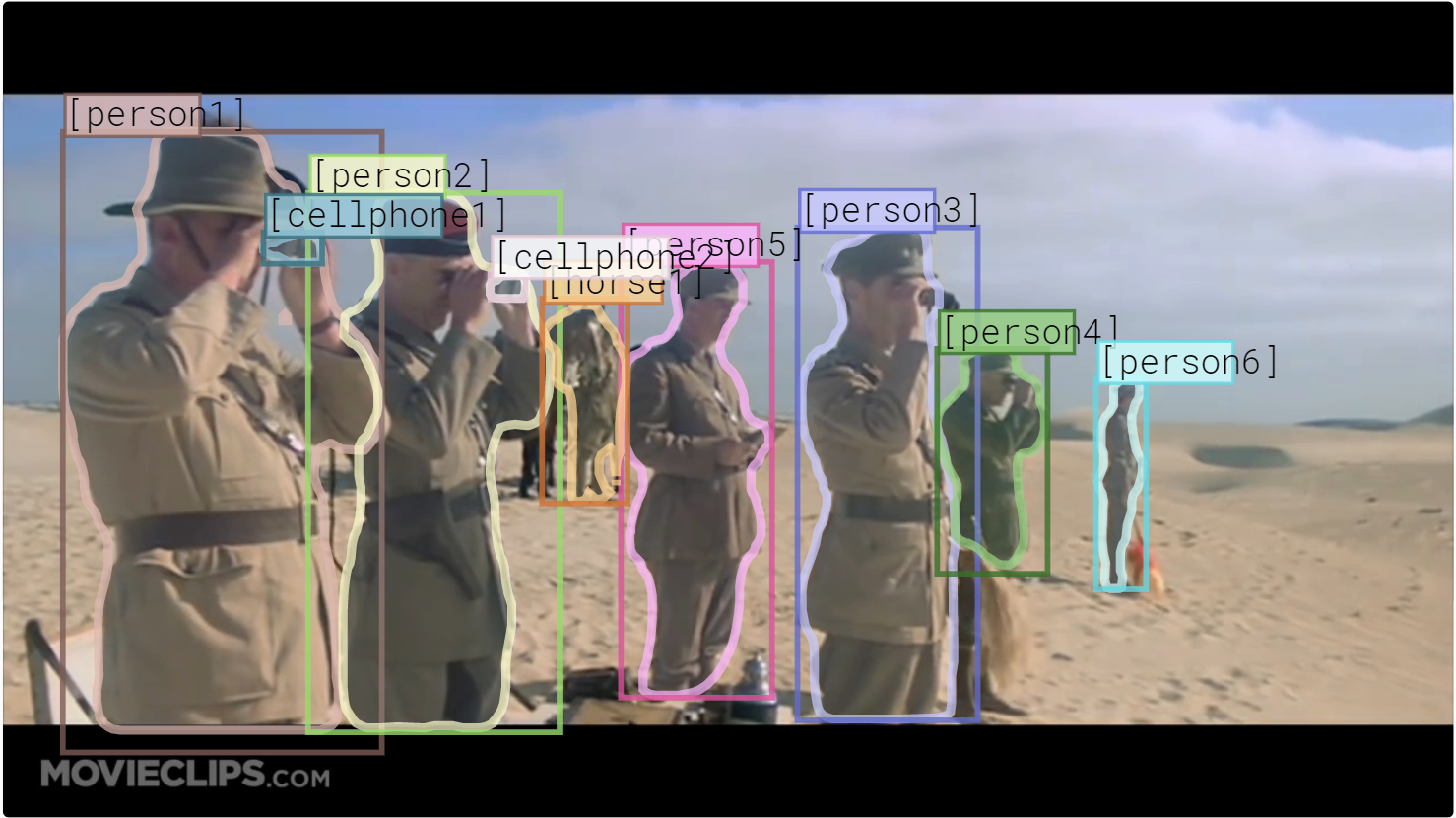}
    \label{fig:image}
  \end{minipage}%
  \begin{minipage}[b]{0.6\linewidth}
  \small
    \centering
    \begin{tabular}[b]{lp{65mm}}
      \toprule
      \textbf{Type} & \textbf{Text}\\
      \midrule
      \textbf{Question} & What are the occupations of [person1], [person2], [person3], and [person4] ? \\
      \textbf{Answer} &  [person1], [person2] and [person3] , and [person4] are among the others military officers. \\
      \textbf{Gold Rationale} & They are all \textcolor{gray}{\hl{wearing}} decorated \textcolor{gray}{\hl{military uniforms}}.\\\hline  
      \textbf{ViLBERT-Fr} & Avon is about to cross the range to save Osiris\\\hline
      \textbf{ViLBERT-Ra} & Myrl Tommie are \textcolor{gray}{\hl{wearing military uniforms}}.\\\hline
      \bottomrule
    \end{tabular}
  \end{minipage}
\caption{Generated rationales samples. Highlighted portions show key points of the rationale for the given answer.}
\label{table:example}
\end{table*}
\vspace{-1em}
\subsection{Results}
\subsubsection{Multi-task objective}
Since we are dealing with multiple losses (and objectives) of answer prediction and rationale generation (\cref{eq:losses}), we report a comparative study on different losses we explored in \Cref{tab:vqa_results}. 

\textit{Weighted Losses}: We vary the $\lambda$ in \cref{eq:losses} as 1, 3, 10 and 1000.

\textit{Uncertainty Loss (var)}: We weight the losses $\mathcal{L_A}$ and $\mathcal{L_R}$ by considering the homoscedastic uncertainty of each task as in \citet{Cipolla_2018}.

\textit{KL-Divergence (kldiv)}: We add Kullback–Leibler divergence \cite{kullback1951} loss between predicted answer scores \cref{eq:soft_scores} and answer scores from pretrained ViLBERT as an added regularizer. This was done to prevent our model from diverging too much from the trained ViLBERT on the original VQA task.

We see from \Cref{tab:vqa_results} that the model trained with KL-Divergence loss performs best on both the VQA task and rationale generation task. As such, we do all further comparison with this model and name it ViLBERT-Ra (ViLBERT-Rationale).

\subsubsection{Quantitative Results}
\textit{Rationale generation: }We compare the performance of ViLBERT-Fr and our model, ViLBERT-Ra in \Cref{tab:results}. We see that our model consistently out performs ViLBERT-Fr over both n-gram metrics -- BLEU and ROUGE and semantic similarity measurement metric -- cosine similarity. This demonstrates how we can leverage rationale generation task to improve model comprehension abilities of existing VQA models.

\textit{VQA task: }We also compare the performance of the two models on the original VQA task in \Cref{tab:vqa_results}.
Since we trained all our models with a batch size of 32 (due to limited compute resources), 
we ran a control experiment to train ViLBERT with the same batch size instead of original 64 for fair comparison. 
We find that ViLBERT-Ra gives superior performance on the rationale generation task without compromising accuracy on the original VQA task (\Cref{tab:results}). In fact, VQA performance is slightly improved. This suggests that training the model to generate rationales can improve model's comprehension which in turn can lead to better answer prediction judgement.
\subsubsection{Qualitative Results}
\textit{Human Evaluation:} We presented 100 randomly selected samples from the VCR validation set containing an image, question, correct answer and rationales generated by ViLBERT-Fr and ViLBERT-Ra to human evaluators. The generated rationales were shuffled randomly to hide which rationale came from which model. We then asked them to choose which of the two candidate rationales better explains the answer in the given sample. The results are summarized in \Cref{tab:human_eval}. Humans consistently rate the rationales generated by ViLBERT-Ra as better explanations for the answers.

We show an illustrative example of a rationale generated by the two models in \Cref{table:example}. During training, we replaced tags like [person1], [person2] with random names, so in both rationales we can observe random names being generated. However, we note that our model was able to generate key words of the gold rationale and convey the relevant meaning. We have provided more such examples in the attached appendix.  
\begin{table}[t]
\small
\centering
  \begin{tabular}{lccccc}
    \toprule
    \textbf{Models} & \textbf{H-1} & \textbf{H-2} & \textbf{H-3} & \textbf{Majority Voting} \\
    \midrule
    ViLBERT-Fr & 31  & 29 & 38 & 20 \\
    ViLBERT-Ra & \textbf{69}  & \textbf{71} & \textbf{62} & \textbf{80} \\
    \bottomrule
  \end{tabular}
  \caption{Percent of rationales preferred by three human judges. Here, H stands for Human.}
  \label{tab:human_eval}
\end{table}

\section{Related Work}

\paragraph{Evaluating VQA model comprehension.}

High performance of VQA models \cite{jiang2018pythia} have naturally led to calls for investigating biases \cite{manjunatha2018explicit} and interpretability.
Researchers have employed various kinds of attention mechanisms over words and images to point out sections of images and words that the model attends to while answering the question \cite{Das_2017, goyal2016transparent, Agrawal_2016_analyze_VQA}.
Another set of approaches use various kinds of 
`selection' tasks
as a means to interpret models. \citet{Goyal_2017} propose picking another image for the same question that has a different answer. \citet{Berg:2013:YTB:2586117.2587016} propose selecting visual facts (image regions) from the image while \citet{zellers2018recognition} propose picking one rationale from a set of four options.

An orthogonal set of methods \cite{Andreas_2016, Hu_2017, Mascharka_2018, vedantam2019probabilistic} approach this task by generating symbolic programs to reason about the question.  We note that such an approach would quickly become intractable given the size of the symbol vocabulary required to cover free-form rationale generation, as we are considering. 




\paragraph{Generating rationales.} The task of generating explanations has previously been employed by \citet{Hendricks_2016} to explain fine-grained bird recognition decisions. In our case, we explain  answers to visual questions with rationales. 
\citet{Rajani_2019} generate reasons and rationales to explain question answering tasks, but only in a purely textual mode.

To the best of our knowledge no prior work has proposed the task of generating rationales as a measure of evaluating comprehensive understanding of images, questions and answers in VQA models.

\section{Conclusion}
In this paper, we proposed the novel task of generating rationales as a measure of model understanding for Visual Question Answering tasks. A well-reasoned explanation implies a thorough understanding of all components of the task: the image, the question and the answer. We further proposed an end-to-end training method to improve the model's commonsense understanding. We demonstrated the effectiveness of our proposed method through 
quantitative and qualitative results.

\bibliography{acl2020}
\bibliographystyle{acl_natbib}




\end{document}